\documentclass[10pt,twocolumn,letterpaper]{article}

\usepackage{wacv}
\usepackage{times}
\usepackage{epsfig}
\usepackage{graphicx}
\usepackage{amsmath}
\usepackage{amssymb}
\usepackage{multirow}
\usepackage{multicol}
\usepackage{bm}

\usepackage{tabularx,booktabs}
\newcolumntype{Y}{>{\centering\arraybackslash}X}
\DeclareMathOperator*{\argmax}{argmax}
\DeclareMathOperator*{\argmin}{argmin}

\newcommand*{\affaddr}[1]{#1} 
\newcommand*{\affmark}[1][*]{\textsuperscript{#1}}
\newcommand*{\email}[1]{\texttt{#1}}

\def\wacvPaperID{202} 

\wacvfinalcopy 
\ifwacvfinal
\def\assignedStartPage{9876} 
\fi

\usepackage[breaklinks=true,bookmarks=false]{hyperref}

\usepackage[dvipsnames]{xcolor}

\ifwacvfinal
\else
\fi

\begin{document}

\title{Towards Visually Explaining Video Understanding Networks with Perturbation}

\author{Zhenqiang Li\affmark[1,2],~ Weimin Wang\affmark[2]\thanks{Corresponding author.},~ Zuoyue Li\affmark[3],~ Yifei Huang\affmark[1],~ and~ Yoichi Sato\affmark[1]\\
\affaddr{\affmark[1]Institute of Industrial Science, The University of Tokyo~~\\ \affmark[2]National Institute of Advanced Industrial Science and Technology~~ \affmark[3]ETH Z\"urich}\\
\email{\{lzq,hyf,ysato\}@iis.u-tokyo.ac.jp}\\
\email{\{li.zhenqiang,weimin.wang\}@aist.go.jp}\\
\email{li.zuoyue@inf.ethz.ch}
}

\maketitle

\begin{abstract}

``Making black box models explainable'' is a vital problem that accompanies the development of deep learning networks. For networks taking visual information as input, one basic but challenging explanation method is to identify and visualize the input pixels/regions that dominate the network's prediction. However, most existing works focus on explaining networks taking a single image as input and do not consider the temporal relationship that exists in videos. Providing an easy-to-use visual explanation method that is applicable to diversified structures of video understanding networks still remains an open challenge. In this paper, we investigate a generic perturbation-based method for visually explaining video understanding networks. Besides, we propose a novel loss function to enhance the method by constraining the smoothness of its results in both spatial and temporal dimensions. The method enables the comparison of explanation results between different network structures to become possible and can also avoid generating the pathological adversarial explanations for video inputs. Experimental comparison results verified the effectiveness of our method.
   
\end{abstract}

\section{Introduction}\label{sec:introduction}
Deep neural networks have achieved remarkable performance in various tasks~\cite{two_stream,i3d,tsn,trn,dense_videocaption,vqa,Huang_2018_ECCV,Huang_2020_CVPR}. Besides good results, sometimes people may also naturally concern about the interpretability of a network, \ie, why a certain prediction is derived by a network for a given input. As one direction towards interpreting these networks, visual explanation (also known as \textit{attribution}) methods, which identify and visualize the contribution of each pixel/region of a given input to the output of a trained network, have attracted much attention recently~\cite{zeiler2014visual, guided_bp_iclr, integrate_grad_icml, lrp, grad_cam, excit_bp_rnn}. 

\begin{figure}[t]
\centerline{\includegraphics[width=1.0\linewidth]{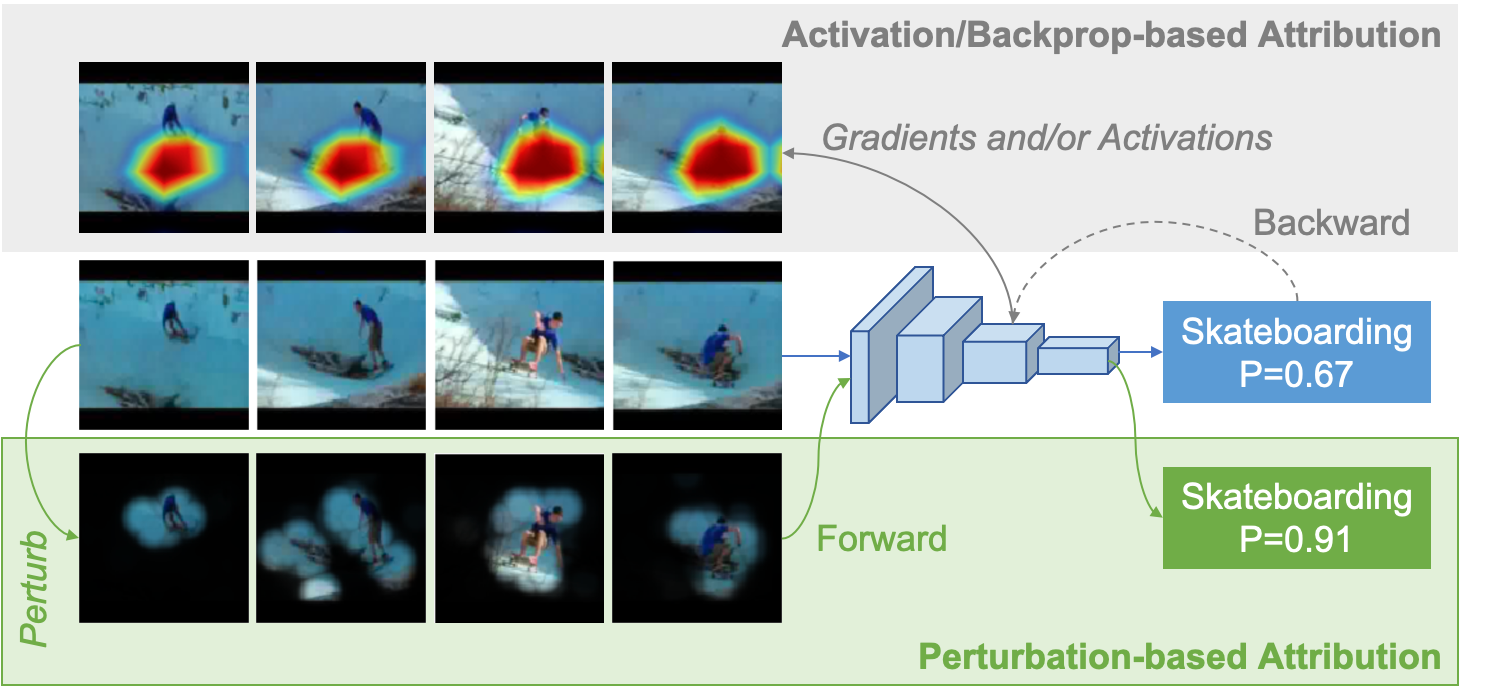}}
\caption{The visual explanation methods for deep neural networks could be generally divided into three categories. The upper block demonstrates the activation-based and backprop-based methods, which usually utilize activations or gradients extracted from the interior layers of a network to identify the significant places in the input frames. The lower block demonstrates the perturbation-based methods, who visually explain a black-box network by directly operating on the input and locating the area that affects the output most in a forward manner. In this paper, aiming at visually explaining video understanding networks via a model-agnostic method, we investigate the perturbation-based attribution method on video classification networks.}
\label{fig:intro}
\end{figure}

Recently, researchers begin to focus on the visual explanation methods of video understanding networks. Most existing works on visual explanation methods are concentrated on individual images~\cite{integrate_grad_icml, smoothgrad, excitbp, rise, grad_cam, meaningful_perturb, extremal_perturb}. Directly applying these methods to videos usually cannot obtain satisfactory explanation results since it is difficult for these methods to handle the complex nonlinear and even recurrent spatiotemporal dependencies in videos. Although there are a few works~\cite{excit_bp_rnn, saliency_tubes} turning on the visual explanation methods for video processing networks, they are designed specifically for only a fixed type of network (\eg 3D-CNNs or CNN-RNN) and cannot be generalized to other networks. 

The perturbation-based approach~\cite{meaningful_perturb, extremal_perturb} is a promising direction for visually explaining video understanding networks since it is agnostic to the network structure. As illustrated in Fig.~\ref{fig:intro}, the perturbation-based method operates on the input and then observes changes in the model outputs. Through iteratively adjusting preserved pixels/regions in the input video and observing the effect on the output of the network, the perturbation-based methods aim to find a small subset of the input, with the preservation of which a large output value can be still retained. However, one cannot directly apply the previous perturbation-based approaches (\eg ~\cite{meaningful_perturb,extremal_perturb}) that were applied to individual images to video understanding networks. This is because pathological explanation results caused by the adversarial effect would be easily produced if no smoothness constraint on the temporal dimension is incorporated~\cite{adversarial,adversarial_intriguing,meaningful_perturb,extremal_perturb}.

In this paper, we propose a generic method for visually explaining video understanding networks by incorporating a perturbation-based method enhanced by a spatiotemporal smoothness loss function. This method can be easily applied to any video understanding networks without detailed architectural knowledge. Furthermore, the loss function exploits the spatiotemporal dependencies between frames to generate explanation results smoothed in both temporal and spatial dimensions and thus avoid the pathological adversarial explanations for video understanding networks. The contribution of this paper is three-fold:
\begin{itemize}
    \item We are the first to introduce perturbation-based method to visually explain video understanding networks. The proposed method can be easily utilized on any diversified and complicated video understanding networks. 
    \item We introduced a novel loss function to regularize the spatiotemporal smoothness of the visual explanation results derived by the perturbation-based method.
    \item Our experimental results verified that the proposed perturbation-based method equipped with our loss function could achieve competitive performances on multiple datasets.
\end{itemize}

\section{Related Work}\label{sec:related_work}
\label{sec:related_work}
In this section, we introduce existing attribution approaches for the visual explanation, including methods mainly focused on networks for individual images (referred to as `image attribution method' below), as well as methods especially proposed for video understanding networks (noted as `video attribution method' below).

\subsection{Image Attribution Approaches}
The goal of an image attribution method is to tell us which elements of the input (\eg, pixels or regions for an image input) are responsible for its output (\eg, the softmax probability for a target label in the image classification task). The results are commonly expressed as an \textit{importance map} in which each scalar quantifies the contribution of the element in the corresponding position. Differing from the attention mechanism that is commonly embedded in a deep network to enhance performance by removing redundancies in input, attribution methods are applied to a model with fixed parameters to provide explanations.

\paragraph{Backpropagation-based (BP-based) methods}
\hspace{-10pt} are established upon a common view that gradients (of the output with respect to the input) could highlight key regions in the input since they characterize how much variation would be triggered on the output by a tiny change on the input. \cite{baehrens2010explain} and \cite{simonyan2013deep} have shown the correlation between the pixels' importance and their gradients for a target label. However, the importance map generated by raw gradients is typically visually noisy. The way to overcome this problem could be partitioned into three branches. DeConvNets \cite{zeiler2014visual} and Guided Backprop \cite{guided_bp_iclr} modify the gradient of the ReLU function by discarding negative values during the back-propagation calculation. Integrated Gradients \cite{integrate_grad_icml} and SmoothGrad \cite{smoothgrad} resist noises by accumulating gradients. However, methods based on unmodified backpropagation tend to capture the average properties of the network, and thus are difficult to obtain class-discriminative explanation results. LRP \cite{lrp}, DeepLift \cite{deeplift} and Excitation Backprop \cite{excitbp} propose modified backpropagation schemes to overcome this challenge. But a modified backpropagation scheme may also limit the user-friendliness of the method, because not all operations used in networks may be compatible by the scheme.

\paragraph{Activation-based methods}
\hspace{-10pt} generate the importance map by linearly combining the activation maps taken from the intermediate convolutional layer of a network. Different methods vary in the choice of combining weights. CAM \cite{cam} selects parameters on the fully-connected layer as weights, while Grad-CAM \cite{grad_cam} produces the weight by average pooling the gradients from the output to the activation. Grad-CAM++ \cite{grad_cam_plusplus} replaces the average pooling in Grad-CAM with coefficients calculated by second derivative. Since the intermediate activation maps are likely to have a lower resolution than input images on networks with pooling layers, the visual explanation results derived by these methods tend to be coarse-grained. For example, when these methods are utilized to visually explain 3D-CNNs with temporal pooling layers, adjacent input frames will be allocated with the same visual explanation result. 

\paragraph{Perturbation-based methods}
\hspace{-10pt} start from an intuitive assumption that the change of outputs could reflect the importance of certain elements when they are removed or preserved only in the input. However, in order to find the optimal results, theoretically it is necessary to traverse the elements and their possible combinations in the input and observe their impact on the output. Due to the high time cost of this traversal process, how to obtain an approximate optimal solution faster is the research focus of this problem. Occlusion \cite{zeiler2014visual} and RISE \cite{rise} perturb an image by sliding a grey patch or randomly combining occlusion patches, respectively, and then use changes in the output as weights to sum different patch patterns. LIME \cite{lime} approximates networks into linear models and uses a super-pixel based occlusion strategy. Meaningful perturbation \cite{meaningful_perturb} converts the problem to an optimization task of finding a preservation mask that can maximize the output probability under the constraints of area ratio and smoothness. Real-time saliency \cite{realtime_saliency} learns to predict a perturbation mask with a second neural network. Qi \etal~\cite{qi2019visualizing} improved the optimization process by introducing integrated gradients and Wagner \etal~\cite{wagner2019interpretable} introduced certain restrictions in the optimization process to avoid adversarial results. Fong \etal~\cite{extremal_perturb} introduced the extremal perturbation scheme and a special smooth mask to solve the problem of imbalance between several constraining terms.

\subsection{Video Attribution Approaches}
The goal of the video attribution is to obtain the regions taken important by a network of the input, in both spatial and temporal dimensions. The increase of dimension means inflated searching space and time cost. \cite{devnet} and \cite{lrp_video} respectively applied pure gradients and LRP to ground the input regions taken important by a video understanding network. However, directly utilizing image attribution methods to videos is likely to obtain unsatisfactory explanation results since spatiotemporal dependencies between frames are not considered by these methods. EB-R (excitation backprop for RNNs) \cite{excit_bp_rnn} firstly extended the Excitation Backprop attribution method to the framework for videos, to be specific, the CNN-RNN structure. Grad-CAM \cite{grad_cam} is inherently applicable to network processing videos. \cite{saliency_tubes} and \cite{stergiou2019class} adapt activation-based methods for 3D convolutional networks to produce visualization results over time. However, both the EB-R and Grad-CAM family cannot treat the network totally as a real black-box, since they have to take the gradients or activations from the interior of a network, which obstructs their application on increasingly complicated and diversified video understanding networks.

\section{Proposed Approach}\label{sec:approach}
In this section, we present the perturbation-based visual explanation method for video understanding networks. Let $\bm{X}=\{\bm{x}_t\}_{t=1}^{T}, \bm{x}_t\in\mathbb{R}^{H\times W\times 3}$ represents a video of $T$ frames with width $W$ and height $H$. The proposed method is investigated on a function $\mathit{\Phi}$ that maps the image sequence to a softmax probability $\mathit{\Phi}_{c}(\bm{x})\in\mathbb{R}$ for a given target class with the index of $c$ among all $C$ classes. The goal of video attribution methods is to derive a sequence of importance maps $\bm{M}=\left\{\bm{m}_t\right\}_{t=1}^{T}$ which assign to each pixel $x_{i,j,t}$ a value $m_{i,j,t} \in \left[0,1\right]$. Here $i,j$ refer to the spatial location of each pixel.

\subsection{Perturbation-based visual explanation}
The \textit{preservation} version of the perturbation-based attribution method~\cite{meaningful_perturb} is to find a reserving subset of the input which is as small as possible while retaining the prediction accuracy on a specified target label. The optimization target can be formulated as follows when the method is applied to a network taking an image $\bm{x}$ as input.

\begin{equation}
    \label{eqn:ptb_ori}
    \bm{m}^*=\argmin_{\bm{m}}\space\{ \lambda||\bm{m}||_{1} - \mathit{\Phi}_c(\bm{m}\otimes{\bm{x}}) \},
\end{equation}
where $\bm{m}$ is the perturbation mask which has the same shape as the input image $\bm{x}$, $||\cdot||_{1}$ denotes the $L_1$ norm for matrices, $\lambda$ controls the scope of regularization, and $\otimes$ represents the local perturbation operation on the input image according to the mask. The operation can be mathematically written as $\bm{m}\otimes{\bm{x}}=\bm{m}\cdot\bm{x}+(1-\bm{m})\cdot(k * \bm{x})$, where $\cdot$ denotes the Hadamard multiplication, $*$ represents convolution, and $k$ denotes a kernel for Gaussian blur. The first item in Eq.~\ref{eqn:ptb_ori} constrains the preservation ratio on the input image to be small while the second item encourages the model's prediction accuracy to be as high as possible.

However, the balance between the two constraint targets is difficult to control, which thus makes it difficult to obtain an optimal solution. In order to alleviate these two negative impacts, we take advantage of the extremal perturbation (\textbf{EP}) according to \cite{extremal_perturb}, which decomposes the optimization procedure into two steps. The first step finds a mask that maximizes the output probability under a constrained preservation ratio $a$, \ie,
\begin{equation}
    \label{eqn:extm_tgt}
    \bm{m}_a=\argmax_{\bm{m}: \parallel\bm{m}\parallel_1=aHW} \mathit{\Phi}_c(\bm{m}\otimes{\bm{x}}).
\end{equation}
The second step sets the lowest bound $\mathit{\Phi}_{0}$ for the output probability, and searches for the smallest mask achieving this bound, \ie, finds the smallest preservation ratio $a^*$ as
\begin{equation}
    a^*=\min\space\{a:\mathit{\Phi}_c(\bm{m}_a\otimes\bm{x})\geq\mathit{\Phi}_{0}\}.
\end{equation}
The final extremal solution $\bm{m}_{a^*}$ is therefore obtained.

Eq.~\ref{eqn:extm_tgt}'s optimization is commonly solved by stochastic gradient descent (SGD) method.
In order to constrain the masks' preservation ratio to approach the setting target $a$, Eq.~\ref{eqn:extm_tgt} is also adjusted by a loss function that regularizes the mask, which enforces values ranked in the top $a$ to be close to 1 and the remaining values to be close to 0, \ie,
\begin{equation}
    \label{eqn:sgd_tgt}
    \bm{m}_a=\argmin_{\bm{m}}\space\{\lambda||\text{vecsort}(\bm{m})-\bm{r}_a||^2-\mathit{\Phi}_c(\bm{m}\otimes\bm{x})\},
\end{equation}
where $\text{vecsort}(\bm{m})\in\mathbb{R}^{HW}$ is a vector in which all the values of a mask are sorted and arranged in a descending order and $\bm{r}_a$ is a template vector consisting of $aHW$ ones followed by $(1-a)HW$ zeros.

\subsection{Spatio-temporal perturbations for videos}
The preservation ratio, which determines how many pixels will be preserved in the input, is a key constraint in the optimization procedure of extremal perturbation. In this paper, we firstly extend the original spatial constraints to the spatiotemporal dimensions, so we can constrain the overall preservation ratio on all the frames. Furthermore, in order to obtain the result with better spatiotemporal smoothness, we introduce an effective loss function, which can consider the inter-frame relationship in the optimization.

\subsubsection{Extremal perturbation 3D}
When adopting extremal perturbations to video cases, we constrain the overall preservation ratio in the whole spatiotemporal dimensions, \ie, the distribution of all masks' values is optimized together in a 3D space. We refer to this method as Extremal Perturbation 3D (\textbf{EP-3D}). Hence, the energy function for optimizing masks under a preservation ratio constraint of $v$ could be represented as
\begin{equation}
    \label{eqn:ep_3d}
    \bm{M}_v=\argmax_{\bm{M}: \sum_{t=1}^T\parallel\textbf{m}_t\parallel_1=vTHW} \mathit{\Phi}_c(\bm{M}\otimes\bm{X}).
\end{equation}
Then we can transfer Eq.~\ref{eqn:sgd_tgt} as below for optimizing by SGD.
\begin{equation}
    \label{eqn:ep_3d_sgd_tgt}
    \bm{M}_v=\argmin\{\lambda||\text{vecsort}(\bm{M})-\bm{r}_v||^2\\-\mathit{\Phi}_c(\bm{M}\otimes\bm{X})\},
\end{equation}
where $\bm{r}_v$ is a template vector consisting of $vTHW$ ones followed by $(1-v)THW$ zeros.

\subsubsection{Spatio-temporal smoothness constraint}
Neural networks are vulnerable to adversarial inputs, \eg, images that are unrecognizable to humans may be recognized by networks as some objects with high confidence~\cite{adversarial_unrecognizable}, and images that are modified in a way imperceptible to humans may mislead networks to have totally wrong predictions~\cite{adversarial_noised}. For perturbation-based methods derived from Eq.~\ref{eqn:ptb_ori}, since their optimization targets are similar to that for generating adversarial inputs~\cite{adversarial, generative_adversarial, adversarial_intriguing}, they are prone to producing pathological solutions that can cause the adversarial inputs. The smoothness constraint is a common method to alleviate the generation of pathological adversarial solution. For example, meaningful perturbations~\cite{meaningful_perturb} incorporated an extra optimization item to regularize the shape of perturbation regions in masks, and extremal perturbations~\cite{extremal_perturb} proposed a special up-sampling operator based on 2D transposed convolution to generate masks with spatial smoothness.
However, only smoothness in the spatial dimensions are constrained by these methods. When extending the perturbation-based method to the video input with an extra temporal dimension, it is also necessary to constrain the smoothness of perturbations in the temporal dimension.

One easy way to constrain mask temporally is to use higher-order differences between frames as an energy function, \eg, using second-order differences to control the smoothness. However, it tends to constrain masks to be consistent which will lose the sensitivity to features on different frames. Considering the spatiotemporal dependencies between frames, a better idea is to smooth masks in spatial and temporal dimensions jointly rather than merely in temporal. Hence we introduce a special loss function $L_K$ that aims to gather the high-value pixels in small regions into some pre-defined shape by doing 3D convolution on the importance maps $\bm{M}$. The $K$ in $L_K$ denotes the kernel used in convolution, which has a shape of $(H_K+1) \times (W_K+1) \times (T_K+1)$. The loss function can be defined as below,
\begin{equation}
    L_K(\bm{M}) = ||\text{vecsort}(\bm{M} * K)-\bm{r}_{v^\prime}||^2,
\end{equation}
where $*$ denotes 3D convolution with stride. In the experiment, we use a ellipsoid kernel which defined as $\forall t\in\{0,...,T_K-1\}, i\in\{0,...,H_K-1\}, j\in\{0,...,W_K-1\}$,
\begin{equation}
    k_{t,i,j}=
    \begin{cases}
      0, &\hspace{-5pt} {\scriptstyle(\frac{2t}{T_K-1}-1)^2+(\frac{2i}{H_K-1}-1)^2+(\frac{2j}{W_K-1}-1)^2 >1}; \\
      1, &\hspace{-5pt} \text{otherwise},
    \end{cases}
\end{equation}
\begin{equation}
    K_{i,j,t}=Z^{-1}k_{i,j,t}.
\end{equation}
Here $Z=\sum_{i,j,t}k_{i,j,t}$ is the normalization factor. The loss function regularizes the first $v^\prime$ high values in the convoluted masks $\bm{M}* K$ to be as close as possible to 1. $v^\prime$ denotes the expected proportion of value-one among all values in the convolution masks, which is calculated as $vTHW/Z$ to satisfy the constraint for preservation ratio in masks $\bm{M}$. In experiment, we set the size of the kernel $K$ to be $11\times11\times7$ and convolution stride to be $11$. This will guarantee the high-value pixels of the mask to be concentrative as much as possible, as well as retain the flexibility in shape to adapt to the spatiotemporal variance in the input frame sequence.

We call this method for visually explaining video understanding networks as \textbf{Saptio-Temporal Extremal Perturbation (STEP)} considering that it could get smoothing extremal perturbation results in both spatial and temporal dimensions.

\section{Experiments and Results}\label{sec:experiment}

\subsection{Experiments setting}
Video classification networks are characterized by complicated and various architectures. To experimentally compare different visual explanation methods for video classification networks, we adopt two kinds of representative structures: CNN-RNN and 3D-CNNs. Specifically, we select two exemplar networks, \ie, VGG16LSTM~\cite{excit_bp_rnn} and R(2+1)D~\cite{r2plus1d} under the two model structures, respectively. We validate methods with the two model structures on subsets of two video datasets: UCF101-24 and EPIC-Kitchens since their bounding box annotations are (partially) available.

\textbf{UCF101-24}~\cite{ucf101} is a subset of the UCF101 dataset, containing 3207 videos of 24 classes that are intensively labeled with spatial bounding box annotations of humans performing actions. In our experiment, we trained a VGG16LSTM model and an R(2+1)D model on the UCF101-24 dataset by the training set defined in THUMOS13. To generate importance maps for evaluating different visual explanation methods, we randomly selected 5 videos on each category to form a test set of 120 videos in total.

\textbf{EPIC-Kitchens}~\cite{epic} is a dataset for egocentric video recognition, where 39596 video clips segmented from 432 long videos are provided, along with action and object labels. We choose the top 20 classes with the most number of clips to form the EPIC-Object and EPIC-Action sub-datasets, and randomly selected 5 clips for each class to generate two test sets and used the remaining clips to train models. Bounding boxes for the ground-truth objects in EPIC-Objects are provided in 2fps. On the EPIC-Action task, we connect a randomly selected part of each clip with its adjacent background frame sequence, to form a set for testing the temporal localization performance of attribution methods.

\subsubsection{Model training}
We trained a VGG16LSTM model and an R(2+1)D model on every video classification task. We use the VGG16LSTM model as it is defined in \cite{excit_bp_rnn} and we fine-tune it on each dataset. To avert the gradient vanishing, we block the gradient propagation on hidden states and take the average of softmax probabilities on all-time steps as the final prediction. For the R(2+1)D model, we use R(2+1)D-18 structure~\cite{r2plus1d} and fine-tune the network upon its pre-trained parameters on Kinetics-400. Both in training and testing phases, we sample 16 frames as the input by splitting one video clip into 16 segments and selecting one frame in each split. The classification accuracy for each network on every task's test set is shown in Tab.~\ref{tab:class_acc}. Notably, the accuracy on the UCF101-24 test set is nearly 100\%. We think this is due to the models are pre-trained on the UCF101 datasets.

\begin{table}[!ht]
    \centering
    \caption[]{Top 1 \& 5 classification acc. of test networks}
	\label{tab:class_acc}
    \setlength{\tabcolsep}{2.5pt}
    \begin{tabularx}{\linewidth}{l|*{3}{Y}}
    \toprule
    Acc.     & UCF101-24 & EPIC-Object & EPIC-Action\\
    \midrule
    R(2+1)D  & 1.00 / 1.00 & 0.57 / 0.85 & 0.77 / 0.97 \\
    V16L     & 0.97 / 1.00 & 0.55 / 0.84 & 0.81 / 1.00 \\
    \bottomrule
    \end{tabularx}
\end{table}

\subsubsection{Evaluation metrics}
As most previous works did, we mainly adopt the `Pointing Game' metric to evaluate the effectiveness of different visual explanation approaches. This metric takes advantage of manually annotated bounding boxes to evaluate whether the importance maps generated by an attribution method could locate the ``key'' spatial regions or temporal segments, which is called spatial pointing game (\textbf{S-PT}) and temporal pointing game (\textbf{T-PT}) respectively. We perform the S-PT evaluation on the UCF101-24 and EPIC-Object test sets. Following \cite{excit_bp_rnn}, we set a tolerance radius of 7-pixel, \ie, one hit is recorded if a 7-pixel radial circle around the maximum point in an importance map intersects the ground-truth bounding box. On the EPIC-Action test set, we evaluate methods by the T-PT metric, in which a hit is recorded only when the index of the frame with the highest importance value locates in the ground-truth segment. The metric is measured by the hit rate on all test samples with bounding box annotations.

\subsubsection{Implementation details}
Following \cite{extremal_perturb}, all masks are generated and optimized based on smaller seed masks $\bar{\bm{M}}=\{{\bar{\bm{m}}_t}\}\in\mathbb{R}^{\bar{H}\times \bar{W} \times T}$ and in our experiment we set $H=7\bar{H}$ and $W=7\bar{W}$. The seed masks are then up-sampled by the transposed convolution operation with the 2D smooth max kernel defined in \cite{extremal_perturb}. We report our explanation results under a series of preservation ratio constraints, which is $\{0.02, 0.05, 0.1, 0.2\}$ for R(2+1)D and $\{0.05, 0.1, 0.2, 0.3\}$ for VGG16LSTM. We expect the redundant information could be removed and key regions could be located via small preservation ratios. Empirically, a larger preservation ratio will not arouse a significant increase in the quantitative results.

\subsection{Comparison between 3D-CNNs and CNN-RNN}

\begin{figure}[ht]
    \centering
    \includegraphics[width=1.0\linewidth]{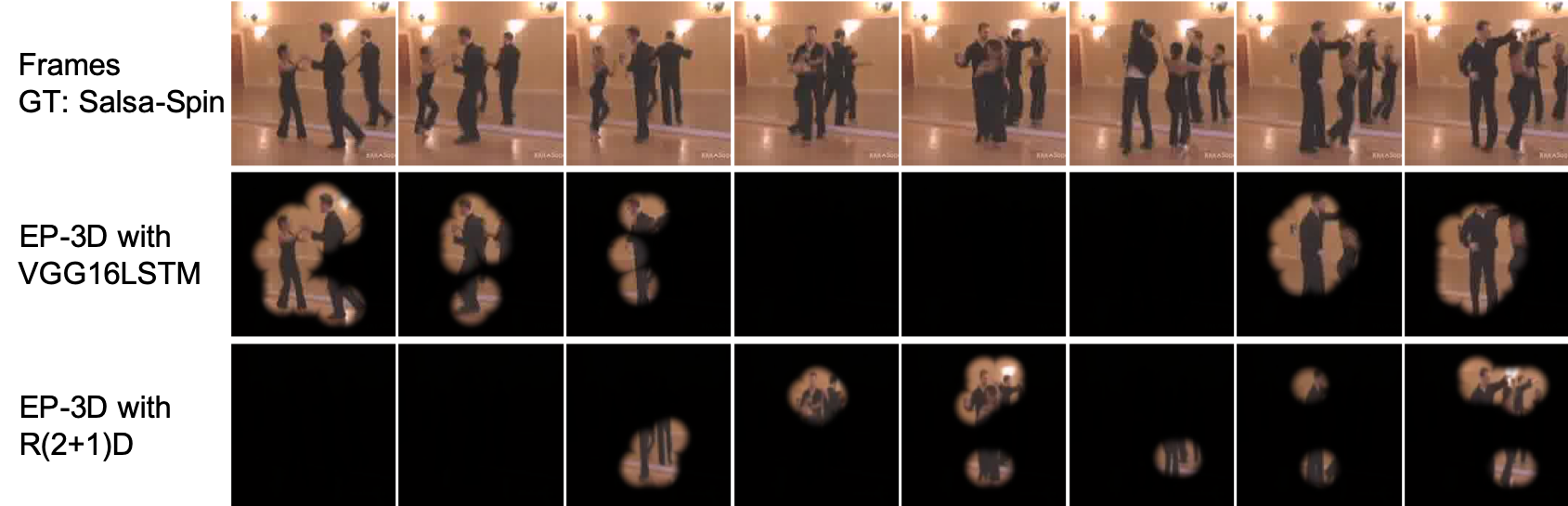}
    \caption{Visual explanation results comparison for two networks. The output probabilities on the two networks are guaranteed to be nearly equal when generating the results by EP-3D.}
    \label{fig:model_comp}
\end{figure}

\begin{table}[ht]
    \centering
	\caption[]{Minimum preservation ratios without decreasing the output probabilities. For the same dataset, VGG16LSTM (V16L) need to preserve more regions than R(2+1)D.}
	\label{tab:smallest_area}
	\setlength{\tabcolsep}{2.5pt}
    \begin{tabularx}{\linewidth}{l|*{2}{Y}}
    \toprule
    Ratio & EPIC-Object & UCF101-24\\
    \midrule
    R(2+1)D    & 0.064 & 0.170 \\ 
    V16L  & 0.157 & 0.280 \\
    \bottomrule
    \end{tabularx}
\end{table}

One of the aims of investigating visual explanation methods is to understand the characteristics of networks, especially when we have multiple networks for the same task. Since we have constructed a generic visual explanation method and two networks for the task of video classification, it is natural to be curious about the difference of explanation results derived through the same method. Fig.~\ref{fig:model_comp} shows the visual comparison between explanation results obtained by EP-3D on R(2+1)D and VGG16LSTM. We uniformly sampled 8 frames for visualization. When obtaining the results, we ensure that perturbed videos have the same probability outputs under each network. It can be seen that VGG16LSTM relies on more input regions than R(2+1)D to derive equal output probabilities. 
For quantitative comparison, we calculated the minimum preservation ratio for each sample in a dataset without decreasing the output probabilities of the two networks and made an average for both datasets. The result, which can be seen in Tab.~\ref{tab:smallest_area}, is consistent with Fig.~\ref{fig:model_comp}, where we found that VG16LSTM (V16L) requires more preservation regions to achieve a comparable output probability as R(2+1)D. We consider that the potential reason for the larger preservation in VGG16LSTM's results relies on its network architecture, where frame features are first extracted by VGG-16 and then forwarded into LSTM recurrently. If the preserved regions are too small in a certain frame, the feature vector for this frame would be close to zero and would become trivial for LSTM. As a result, VGG16LSTM tends to preserve more regions in order for higher prediction accuracy.

\begin{figure*}[ht]
    \centering
    \includegraphics[width=1.0\linewidth]{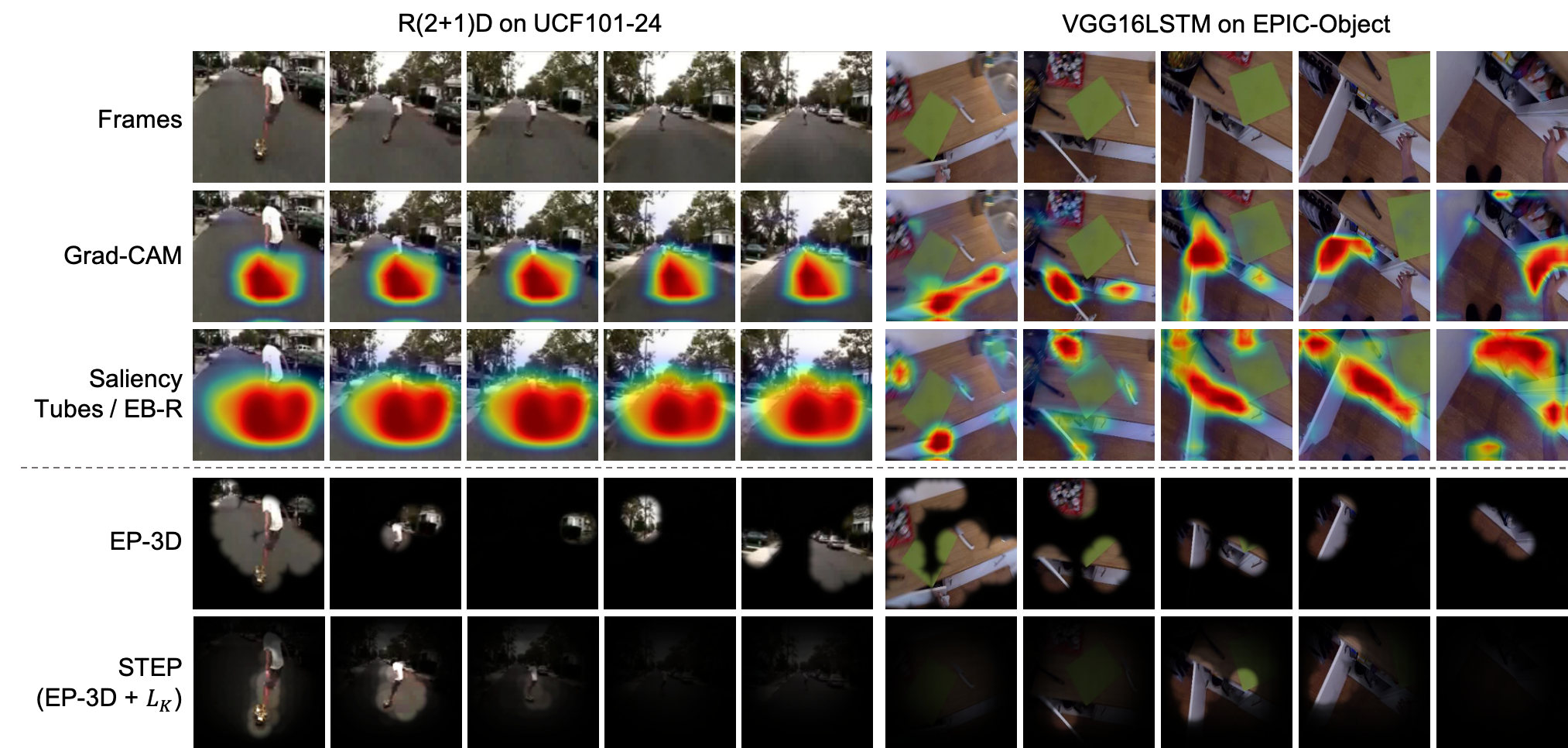}
    \caption{Qualitative comparison of visual explanation results generated by baseline methods and the perturbation-based methods. The importance maps generated by our method could smoothly preserve the regions associated with the ground-truth label and remove areas with weaker correlations.}
    \label{fig:vis_res_comp}
\end{figure*}

\subsection{Comparison of existing attribution methods}
We compare with three existing video attribution methods as baseline methods to validate the effectiveness of our proposed method.
\begin{itemize}
    \item \textbf{Grad-CAM\cite{grad_cam}}: A generic attribution method that \textit{could be utilized on both 3D-CNNs and CNN-RNN networks}. For the R(2+1)D model, we generate the heatmaps based on the activation of the last 3d convolutional layer and upsample the maps to the shape of input images, in both spatial and temporal dimension. For the VGG16LSTM model, the heatmaps are generated based on the activation of $conv5$-layer of VGG16.
    
    \item \textbf{Saliency Tubes\cite{saliency_tubes}}: A visualization method \textit{specially designed for 3D-CNNs networks}. The activation maps of the last 3d convolutional layer are combined by weights in the final FC layer to produce heatmaps. We up-sample the maps as in Grad-CAM for visualization and evaluation.
    
    \item \textbf{EB-R\cite{excit_bp_rnn}}: A backprop-based method \textit{specially proposed for the CNN-RNN structure} which uses a modified back-propagation algorithm. We adopt it directly on our VGG16LSTM models and capture the heatmaps for each frames at the $conv5$ layer of VGG16 as it was done in \cite{excit_bp_rnn}.
\end{itemize}

\subsubsection{Qualitative results}
Fig.~\ref{fig:vis_res_comp} illustrates two groups of visualization results including the original frames and importance maps generated by different visual explanation approaches. Each group corresponds to one example video, and only 5 frames are sampled out of 16 input frames for visualization. For perturbation-based methods, we visualize the results generated under the preservation ratio constrain of 0.1. The first group shows a UCF101-24 video with the action label of Skateboarding and all visual explanation results are obtained on the R(2+1)D model. A video with the object label of Cupboard belonging to the EPIC-Object sub-dataset is showed as the second group, whose visual explanation results are all obtained on the VGG16LSTM.

There is a part of regions highly correlated with the ground-truth label in the two videos, \eg, the first two frames of the left video example, and the middle three frames of the right example. STEP enhances EP-3D by a special loss function. It can be seen that STEP could generate temporally smooth mask sequences to preserve these related regions. In contrast, the regions preserved by the mask sequences generated by EP-3D perform lower consistency in the temporal dimension. They also tend to contain some regions showing no obvious relationship with the ground-truth label, such as the region of a red object in the second frame of the cupboard video. This reflects the effectiveness of $L_K$ in improving the temporal smoothness of preservation masks. Moreover, combining with comprehensive observations of other samples' results, we discover that EP-3D tends to allocate more preserved regions on the head and tail frames for R(2+1)D and the front frames for VGG16LSTM. 

On the R(2+1)D model, although stable results could be generated by Grad-CAM, they are coarse-grained in both spatial and temporal dimensions, which is the same for Saliency Tubes. This is because the two methods both generate importance maps based on the activation maps extracted from the intermediate convolutional layer, which have a lower resolution than the input frame sequence when the pooling operators are applied. On VGG16LSTM, EB-R could locate the ground-truth object while it also highlights the salient but unrelated regions.

\subsubsection{Quantitative results}
We then quantitatively compare the visual explanation results generated by different methods, using the spatial pointing game (S-PT) metric, which measures the percentage of importance maps whose maximum points fall into the annotation bounding boxes. When calculating the metric, only the frame annotated with bounding boxes are considered. The evaluation results on the two models and two datasets are shown in Tab.~\ref{tab:quan_spg}. It can be seen that in all cases, STEP is able to get the best performance and achieve obvious improvement on EP-3D. A potential reason could be that the proposed loss function $L_K$ smooths the perturbations in both spatial and temporal dimensions so that the regions unrelated with the specified label are removed.

\begin{table}[ht]
    \centering
	\caption[]{Quantitative evaluation of visual explanation results generated by different method based on the spatial pointing game (S-PT) metric (in percentage). Results on this metric are measured by percentage. Here we use V16L to represent VGG16LSTM for short.}
	\label{tab:quan_spg}
    \begin{tabularx}{\linewidth}{l *{4}{Y}}
    \toprule
    Methods
    & \multicolumn{2}{c}{EPIC-Object}  
    & \multicolumn{2}{c}{UCF101-24}\\
    \cmidrule(lr){2-3} \cmidrule(l){4-5}
    & R(2+1)D & V16L & R(2+1)D & V16L \\
    \midrule
    Grad-CAM~\cite{grad_cam} & 7.1 & 6.1 & 47.5 & 35.5 \\
    SaliencyTubes~\cite{saliency_tubes} & 6.6 & / & 41.4 & / \\
    EB-R~\cite{excit_bp_rnn} & / & 6.9 & / & 46.5 \\
    \midrule
    Ours (EP-3D) & 8.1 & 6.1 & 47.6 & 47.2 \\ 
    Ours (STEP) & \textbf{9.0} & \textbf{9.1} & \textbf{56.7} & \textbf{52.7} \\
    \bottomrule
    \end{tabularx}
\end{table}

\subsection{Comparison of methods for smoothness}
To further evaluate the effectiveness of our proposed loss function on enhancing the EP-3D method through constraining temporal smoothness, we designed the following baseline smoothness methods for comparison. The quantitative results are available in Tab.~\ref{tab:quan_smooth_baseline}.

\vspace{0.5em}
\textbf{Evenly allocating preservation in temporal} is a straight-forward way, \ie, assigning each frame the same preservation ratio. Thus under this setting, we only optimize the distribution of mask values on the 2D image space of each frame. Formally, if we set the constraint to the overall preservation ratio as $v$, then the energy function Eq.~\ref{eqn:ep_3d} and \ref{eqn:ep_3d_sgd_tgt} for EP-3D could be transcribed as follow in the video case,
\begin{equation}
    \label{eqn:ep_2d}
    \bm{M}_v=\argmax_{\bm{m}_t:||\bm{m}_t||_1=vHW, \forall t} \mathit{\Phi}_c(\bm{M}\otimes\bm{X}),
\end{equation}
\begin{equation}
    \label{eqn:ep_2d_sgd_tgt}
    \bm{M}_v=\argmin_{\bm{M}} \{\lambda\sum_{t=1}^T||\text{vecsort}(\bm{m}_t)-\bm{r}_v||^2-\mathit{\Phi}_c(\bm{M}\otimes\bm{X})\}.
\end{equation}
which means we constrain the preservation ratio for each frame's mask uniformly as the predefined $v$. Here $\bm{r}_v$ is a template vector consisting of $vHW$ ones followed by $(1-v) HW$ zeros. We refer to the variant method of EP-3D as \textbf{EP-3D-Evenly} for short below. We see this method as one straight-forward method for temporal smoothness because the distribution of preservation ratios on frames are uniform. 

It can be seen from Tab.~\ref{tab:quan_smooth_baseline} that EP-3D-Evenly have comparable performance as the pure EP-3D on R(2+1)D models. On VGG16LSTM models, EP-3D-Evenly achieves obvious improvements than the pure EP-3D, which is consistent with our analysis that CNN-RNN tends to give even focuses on each frame.

\vspace{0.5em}
\textbf{Gaussian smoothing} is a generic method for blurring. As shown in equation~\ref{eqn:smooth_baseline}, we exploit a Gaussian kernel $k \in \mathbb{R}^{2\Delta t+1} $ to smooth the value of $\bar{m}_{i,j,t}$ on the smaller seed masks $\bar{\bm{M}}$ according to its neighbours in the temporal dimension and yield the smoothed value $\bar{m}'_{i,j,t}$.
\begin{equation}
    \label{eqn:smooth_baseline}
    \bar{m}'_{i,j,t}=Z^{-1}\sum_{t'=t-\Delta{t}}^{t+\Delta{t}} k_{t'-t} \bar{m}_{i,j,t'}
\end{equation}
Here $Z$ normalizes the kernel to sum to one. The kernel $k$ is a radial basis function with profile $k_u=\exp(-u^2/(0.6\sigma))$ and set $\sigma=\Delta t$ to ensure the kernel's sharpness. Since the smoothness operator is applied on the smaller seed masks $\bar{\bm{M}}$ which will be up-sampled to the masks $\bm{M}$ before perturbing frames, the operation could be viewed as smoothing the masks $\bm{M}$ both spatially and temporally. We insert this smoothness operator to the pure EP-3D method and test its effectiveness by choosing three different values of $\Delta{t}$. As shown in Tab.~\ref{tab:quan_smooth_baseline}, the smoothness operator does not improve the performance of EP-3D obviously compared with our proposed $L_K$. We think this is because the smoothness method based on the loss function $L_K$ could give the masks more searching freedom of shape in the process of optimization.

\begin{table}[ht]
    \centering
	\caption[]{Effectiveness evaluation of the proposed loss function $L_K$. We quantitatively compare it with a baseline temporal smoothness method based on Gaussian kernels. Here $\Delta{t}$ decides the size of the applied Gaussian kernel.}
	\label{tab:quan_smooth_baseline}
	\begin{tabularx}{\linewidth}{l *{4}{Y}}
    \toprule
    Methods
    & \multicolumn{2}{c}{EPIC-Object}  
    & \multicolumn{2}{c}{UCF101-24}\\
    \cmidrule(lr){2-3} \cmidrule(l){4-5}
    & R(2+1)D & V16L & R(2+1)D & V16L \\
    \midrule
    EP-3D & 8.1 & 6.1 & 47.6 & 47.2 \\ 
    EP-3D-Evenly & 7.9 & 8.7 & 47.4 & 53.2 \\
    EP-3D (${\Delta}t=1$) & 8.4 & 7.4 & 48.5 & 51.1 \\ 
    EP-3D (${\Delta}t=2$) & 8.1 & 8.1 & 47.9 & 52.6 \\ 
    EP-3D (${\Delta}t=3$) & 7.7 & 7.1 & 50.3 & \textbf{54.3} \\ 
    \midrule
    STEP & \textbf{9.0} & \textbf{9.1} & \textbf{56.7} & 52.7 \\
    \bottomrule
    \end{tabularx}
\end{table}

\subsection{Temporal pointing game}
In Tab.~\ref{tab:quan_tpg}, we show the results for quantitatively evaluating the ability of different visual explanation methods to locate the key action segments in the temporal dimension. The ability is measured by the metric of the temporal pointing game (T-PT) on the EPIC-Action sub-dataset. Because the temporal resolution of the importance maps generated by Grad-CAM and Saliency Tubes on 3D-CNNs is very low, that is, only two different importance maps are generated for the input sequence with 16 frames, we will not test the two methods on R(2+1)D model. The method EP-3D-Evenly is also excluded because it gives each frame the same importance. It can be seen that EB-R performs well on the VGG16LSTM model and the pure EP-3D does not show satisfactory performance on both networks. As we analyzed, the pure EP-3D tends to attribute the network's output on the head and tail parts of videos, which lowers its performance on temporal localization. This could be effectively alleviated after adding the Gaussian smoothness kernel as the results shown in the middle three lines, which also reflects the necessity of introducing temporal smoothness. STEP could also improve the performance of EP-3D by introducing the loss function $L_K$. The improvement is comparable to the best of that achieved by Gaussian kernel on the R(2+1)D model but is relatively lower than the best results on the VGG16LSTM model. 

\begin{table}[ht]
    \centering
	\caption[]{Quantitative evaluation results on the EPIC-Action test set by the temporal pointing game (T-PT) metric (in percentage). Results on this metric are measured by percentage.}
	\label{tab:quan_tpg}
    \begin{tabularx}{0.75\linewidth}{l *{2}{Y}}
    \toprule
    Methods &  R(2+1)D & V16L \\
    \midrule
    EB-R~\cite{excit_bp_rnn} & / & \textbf{57.0}  \\
    \midrule
    EP-3D & 34.0 & 31.0 \\ 
    EP-3D (${\Delta}t=1$) & 46.0 & 39.0 \\ 
    EP-3D (${\Delta}t=2$) & 46.0 & 47.0 \\ 
    EP-3D (${\Delta}t=3$) & 41.0 & 45.0 \\ 
    \midrule
    STEP & \textbf{47.0} & 49.0 \\ 
    \bottomrule
    \end{tabularx}
\end{table}



\section{Conclusion}
In this paper, we shed light on the task of visually explaining video understanding networks by the perturbation-based method. The proposed method is characterized by model-agnostic and thus could be applied to diverse structures of video understanding networks in the same way even without detailed architectural knowledge. We also introduced a novel loss function to smooth the perturbation results in both spatial and temporal dimensions in order for the spatiotemporal smoothness of the explanation results. We experiment on two typical kinds of video classification network 3D-CNNs \& CNN-RNN and on two datasets EPIC-Kitchens \& UCF101-24, and compare smoothness with naive Gaussian blur. Both qualitative and quantitative results show that our proposed method could achieve competitive performance and the effectiveness of our method is therefore verified. 

\section*{Acknowledgments}
This work is partially supported by JST AIP Accelerated Program Grant Number JPMJCR20U1, a project commissioned by the New Energy and Industrial Technology Development Organization (NEDO) and JSPS KAKENHI Grant Number JP20H04205.

{\small
\bibliographystyle{ieee_fullname}
\bibliography{egbib}
}

\end{document}